\documentclass[a4paper, 10 pt, conference]{ieeeconf}
\IEEEoverridecommandlockouts

\usepackage{cite}
\usepackage{amsmath,amssymb,amsfonts}
\usepackage{algorithmic}
\usepackage{graphicx}
\usepackage{textcomp}
\usepackage{xcolor}
\usepackage{multirow}
\usepackage{stfloats}
\usepackage{booktabs}
\usepackage[left=1.57cm,right=1.57cm,top=0.95cm,bottom=2.54cm]{geometry}
\def\BibTeX{{\rm B\kern-.05em{\sc i\kern-.025em b}\kern-.08em
    T\kern-.1667em\lower.7ex\hbox{E}\kern-.125emX}}

\renewcommand{\baselinestretch}{0.97}

\begin{document}

\title{\LARGE The Social Life of Industrial Arms: How Arousal and Attention Shape Human-Robot Interaction}

\author{\authorblockN{Roy El-Helou} 
\authorblockA{\textit{Ingenuity Labs Research Institute} \\
\textit{Queen's University}\\
Kingston, Canada \\
19reh2@queensu.ca}
\and
\authorblockN{Matthew K.X.J. Pan} 
\authorblockA{\textit{Ingenuity Labs Research Institute} \\
\textit{Queen's University}\\
Kingston, Canada \\
matthew.pan@queensu.ca}
}
\maketitle
\begin{abstract}
This study explores how human perceptions of a non-anthropomorphic robotic manipulator can be shaped by two key dimensions of behaviour: arousal, defined as the robot's movement energy and expressiveness, and attention, defined as the robot’s capacity to selectively orient toward and engage with a user. We present an integrated behaviour system that applies and extends existing movement-centric design principles to non-anthropomorphic robots. Our system combines a gaze-like attention engine with an arousal-modulated motion layer to explore how expressive and interactive behaviours influence social perception in robotic manipulators. In a user study, we find that robots exhibiting high attention—actively directing their focus toward users—are perceived as warmer and more competent, intentional, and lifelike. In contrast, high arousal—characterized by fast, expansive, and energetic motions—increases perceptions of discomfort and disturbance. Importantly, a combination of focused attention and moderate arousal yields the highest ratings of trust and sociability, while excessive arousal diminishes social engagement. These findings offer design insights for endowing non-humanoid robots with expressive, intuitive behaviours that support more natural human-robot interaction.
\end{abstract}

\begin{figure*}[hb!]
    \centering
    \includegraphics[width=1\linewidth]{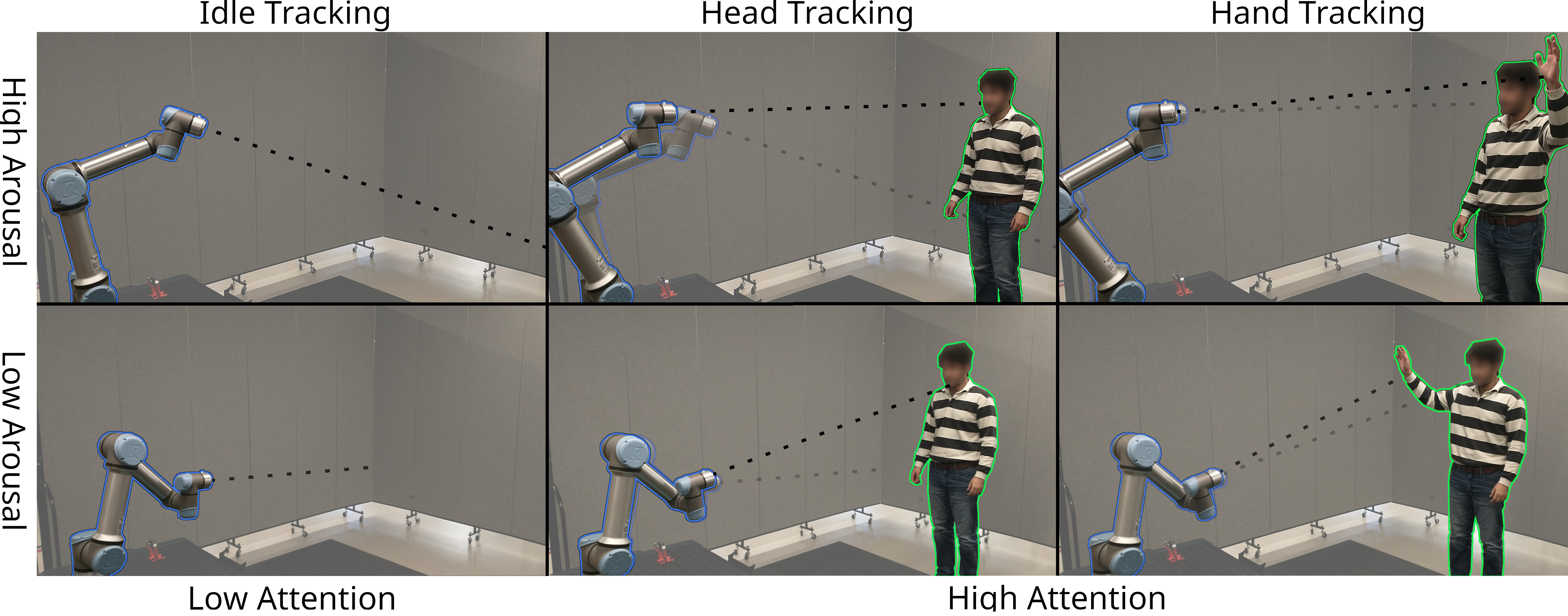}
    \caption{Photos showing varying robot poses as a result of varying the arousal level of the proposed architecture controlling the emotional and social behaviours of a Universal Robots UR5e manipulator. }
    \label{fig:robo-behave}
\end{figure*}

\section{Introduction}
As robots become increasingly integrated into human environments—from homes and classrooms to healthcare and industrial settings—designing them to interact naturally and intuitively with people is becoming a central challenge in human-robot interaction (HRI). While much of robotics has focused on functional performance, there is growing recognition that social and emotional capabilities are equally crucial for fostering trust, collaboration, and long-term engagement \cite{law_interplay_2021}. 

This work explores a fundamental question: Why and how should robots be designed with character-like qualities? We are particularly interested in how robots can use behavioural cues—such as movement, posture, and attention—to project emotional and social presence, even without anthropomorphic features. By endowing robots with the ability to exhibit emotionally expressive behaviour, we aim to deepen human-machine interaction and improve the quality of user experience. In such an endeavour, it is essential to consider not only the capability of robots to communicate emotions but also how humans perceive these expressions. For instance, the findings from \cite{szafir_pay_2012} suggest that adaptive emotional robots can foster rapport and immediacy in educational settings, thus leading to improved learning outcomes. Furthermore, according to \cite{bickmore_establishing_2005}, establishing emotional bonds and providing consistent support are key aspects of long-term human-computer relationships, which can be applied to addressing social isolation and similar challenges. 


In this work, we focus on non-anthropomorphic robots (NARs) —specifically, robotic manipulators—and examine how they can express emotional character through two key behavioural dimensions: \emph{arousal}, defined as the robot’s level of energy or movement intensity, and \emph{attention}, defined as its capacity to selectively orient its ``gaze" toward and engage with human users. Both dimensions are demonstrated in Fig. \ref{fig:robo-behave}. Our proposed system architecture integrates an arousal-modulated motion model, inspired by prior work on expressive movement in robotics, \cite{venture_robot_2019, hoffman_designing_2014, pan_realistic_2020}, with an interactive focus mechanism we call the attention engine. To evaluate this architecture, we conduct a user study examining how variations in arousal and attention affect human perceptions of the robot’s warmth, competence, animacy, and social intentionality. The results offer insights into how expressive motion and attentional behaviours can enhance the social presence of NARs.


\section{Related Work}
\label{rel-wrk}
Traditionally, the ability to convey emotions has been reserved for more anthropomorphic robots –machines designed with human-like features that naturally lend themselves to the expression of human emotions \cite{spatola_ascribing_2021}. Anthropomorphic robots have been widely studied in social human-robot interaction (HRI), particularly for their ability to use both verbal and non-verbal communication modes like kinesics, proxemics, haptics, and voice modulation \cite{chevalier_dialogue_2017, kim_how_2014, green_whos_2022}. 
Studies on such robots emphasize that nonverbal behaviours, including gestures and postures, can shift human cognitive framing, elicit emotional responses, and improve task performance \cite{saunderson_how_2019}. Furthermore, eye gaze plays a critical role in human-robot communication, irrespective of the robot’s form. It has been shown to direct user attention, regulate conversational flow, and convey engagement or intent. Eye gaze has been used in multi-modal interaction to combine attention-directing cues with other communication modes, improving the robot’s ability to integrate into human environments \cite{stanton_robot_2014, moon_meet_2014}. However, anthropomorphic robots also face social hurdles, such as the uncanny valley, a phenomenon in which a certain level of human likeness elicits feelings discomfort \cite{mori_uncanny_2012}.

Instead, our work explores using NARs to behave socially and emotionally.  
NARs, such as robotic manipulators, lack human-like attributes, which makes the task of emotional expression different and more challenging. NARs often appear mechanical and devoid of personal content, lacking the emotional depth observed in their anthropomorphic counterparts \cite{kim_robotic_2022}. However, prior work on social and emotional content designed for NARs discusses how postural adjustments in robots can shift emotional framing and trigger specific human responses \cite{scheflen_significance_1964, erden_emotional_2013, zecca_whole_2009}. Recent work has shown that expressive movement can significantly improve robot readability and engagement; studies incorporating animation principles, such as anticipation and reaction, demonstrate that motion design influences how users perceive a robot’s competence and intelligence \cite{takayama_expressing_2011}. Additionally, a movement-centric approach suggests that robot motion should be designed as a primary communication channel rather than an auxiliary feature \cite{hoffman_designing_2014}. Beyond animation-driven readability, the interplay between expressive and functional movement has been studied in NARs. The ELEGNT framework proposes that robots should integrate both function-driven and expression-driven movements to enhance user engagement and communication, even in robots designed primarily for task execution \cite{hu_elegnt_2025}. This work aligns with research showing that expressive motion helps users anticipate robot actions and infer internal states, making interactions more intuitive. 

\section{Motivations}
While prior work has examined expressive motion and social signalling in robots \cite{breazeal_emotion_2003}, \cite{hoffman_designing_2014}, \cite{takayama_expressing_2011}, few studies have integrated these behaviours into a unified framework for robotic manipulators. Our work builds on this movement-centric perspective and focuses specifically on two core behavioural dimensions: arousal and attention. 

We chose arousal, defined as movement energy or intensity, because it maps directly to robot motion, making it particularly well-suited for non-anthropomorphic platforms that lack facial or vocal expressivity. High arousal behaviours, such as fast, expansive motion, can convey urgency or excitement, while low arousal produces subdued, hesitant, or calm responses \cite{breazeal_emotion_2003}, \cite{erden_emotional_2013}. Unlike valence, as defined by Russell in \cite{russell_circumplex_1980}, which relies on affective expressions difficult to convey without anthropomorphic features, arousal is inherently expressible through kinematics alone. 

Gaze-based attention was selected for its critical role in engagement and social presence. A robot's ability to direct its gaze or orient toward users conveys intentionality and responsiveness, shaping perceptions of agency, sociability, and competence \cite{stanton_robot_2014}, \cite{moon_meet_2014}, \cite{admoni_social_2017}. In non-humanoid robots, attention can be approximated through gaze-like behaviours embedded in motion, allowing the system to remain expressive despite lacking eyes or a face. 

These two dimensions offer complementary yet distinct ways for robotic arms to communicate internal state and interpersonal awareness. We now describe the architecture developed to implement and evaluate these behaviours in a robotic manipulator.

\section{Contributions}
We present a system that integrates and adapts known design principles, such as arousal-modulated motion and interactive attention for non-anthropomorphic robots \cite{venture_robot_2019}, \cite{hoffman_designing_2014}. While the components of this system are grounded in prior work \cite{takayama_expressing_2011}, \cite{pan_realistic_2020}, \cite{hu_elegnt_2025}, our contribution lies in combining them within a unified framework and evaluating their social and emotional impact in a user study. Our contributions are as follows:

\begin{itemize}
    \item We design an expressive control system that integrates arousal-based motion modulation with a real-time attention engine in a robotic manipulator.
    \item We conduct a factorial user study that evaluates how attention and arousal jointly shape perceptions of discomfort and sociability in NARs.
    \item We provide empirical insights into the social dynamics of expressive motion in non-humanoid robots, supporting their use in socially embedded contexts.
\end{itemize}
These contributions are framed around the following research questions:

\begin{enumerate}
    \item How can NARs convey emotional character through movement and interactivity without relying on anthropomorphic features?
    \item What are the key factors in robotic movement and behaviours that can be optimized to evoke specific emotional responses from humans and enhance their emotional engagement? 
    \item How do varying levels of arousal and attention influence human perceptions of a robot’s social presence and intentionality?
\end{enumerate}

With these contributions in mind, we next describe the system architecture developed to explore how arousal and attention shape human perceptions of NARs. This framework provides the basis for answering our research questions through controlled behavioural modulation and interactive experimentation.

\section{System Design}
\label{sec:system_design}
\begin{figure}[bt]
    \centering
    \includegraphics[width=1\linewidth]{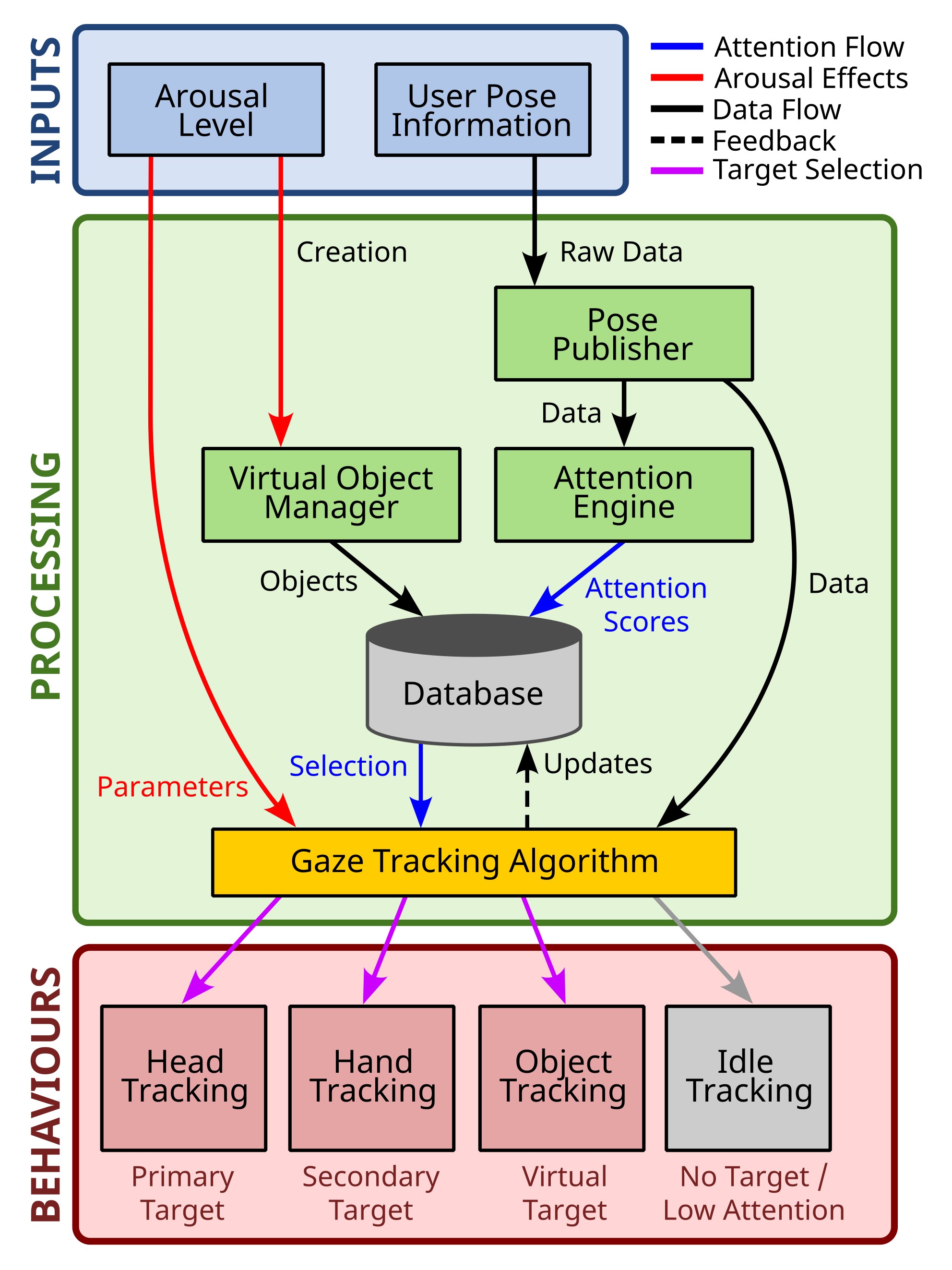}
    \caption{System architecture diagram for our proposed interactive and expressive robot behaviour system. }
    \label{fig:sys-arch}
\end{figure}


Our behaviour generation system integrates attention and arousal mechanisms to support expressive and socially responsive interaction in robotic manipulators. It enables the robot to dynamically orient to human presence while modulation its energy and motion profile to convey emotional character. As shown in Fig.~\ref{fig:sys-arch}, the system consists of two main components: an \textit{attention engine}, which determines which stimuli the robot should attend to (Section~\ref{subsec:attn_eng}), and a \textit{gaze-tracking algorithm} (Section~\ref{subsec:gaze_tracking}). We implemented this system on a Universal Robots UR5e manipulator~\cite{noauthor_ur5e_nodate}, illustrated in Fig.~\ref{fig:robo-behave}, and evaluated it through a user study examining human perceptions of the robot (Section~\ref{sec:user_study}).

\subsection{Attention Engine}
\label{subsec:attn_eng}

As its name suggests, the attention engine is the component in our system which directs the robot's attention to salient features/stimuli in its visual environment; in our case these stimuli are users and individuals within range of a RGB-D sensor (Intel RealSense D435 Depth Camera) affixed in front of the robot \cite{noauthor_intel_nodate}. It builds on an approach introduced by Pan et al.~\cite{pan_realistic_2020}, that was utilized for a humanoid animatronic bust and governs the robot’s decision-making by computing an \textit{attention score} for each detected individual, determining saliency and guiding the robot’s gaze to enable dynamic, responsive interactions.


Each attention score is a weighted combination of positional cues, movement dynamics, and a habituation factor that prevents prolonged fixation on a single target—reflecting how humans distribute attention across multiple people and stimuli. The factors and weights of the score were developed through trial and error testing and subjective feedback from participants of what the robot should do when presented with different stimuli.  

The score $\Phi$ is defined as:

\begin{equation}
    \Phi = w_pP+w_vV+ \Theta(t)
\end{equation}

\noindent where $w_p$ and $w_v$ represent the weight for the position and velocity components of an individual such as their torso and hands in the environment, respectively. The weights are chosen based on their relative importance and influence on attention. $\Theta(t)$ is a habituation factor, described in further detail below. The position score $P$ combines proximity and hand position factors:

\begin{equation}
     P = w_{proximity}e^{\lambda d} + w_{hand}(h_{left} + h_{right})
 \end{equation}
 
\noindent where $d$ is the Euclidean distance to the individual, $\lambda$ is a decay constant, and $h_{\text{left}}$, $h_{\text{right}}$ are binary indicators of whether either hand is raised above shoulder level. This formulation prioritizes nearby individuals and interprets raised hands as social gestures, such as waving. $\lambda$ is used to modulate the influence of distance on the attention score, ensuring that closer individuals are prioritized, similarly to natural human tendency. 

The velocity score $V$ captures torso and hand motion, normalized by maximum observed velocities:

\begin{equation}
    V = \frac{v_{torso}}{v_{max\_torso}} + \frac{v_{right}}{v_{max\_right}} + \frac{v_{left}}{v_{max\_left}}
\end{equation}

\noindent Thus, the faster an individual moves, the more salient (i.e., having a larger attention score) the individual becomes; normalization ensures responsiveness across varying activity levels. 

To avoid sustained attention on a single user, we apply a habituation factor $\Theta(t)$ that dynamically adjusts based on gaze history:

\begin{equation}
    \Theta(t) =\Theta(t-1) + (\gamma m_{hab} + (1 -\gamma)m_{rest})\Delta t
\end{equation}

\noindent Here, $\gamma = 1$ for the currently attended individual and $0$ for others; $m_{\text{hab}}$ and $m_{\text{rest}}$ control the negative decay and positive recovery rates respectively, and $\Delta t$ is the time step. Which provides a natural decay in attention for sustained interactions while allowing recovery during periods of disengagement. $\Theta(t)$ is bounded between 0 and 1 to maintain naturalistic attention shifts and prevent fixation.

\subsection{Robotic Gaze Algorithm}
\label{subsec:gaze_tracking}
The gaze algorithm maps 3D target positions to joint-space postures, enabling the robot to orient toward salient individuals. Though implemented an a UR5e arm, the approach is generalizable to any articulated platform. It prioritizes distal joints to produce localized gaze-like motion without requiring full-body movement. Arousal modulates posture and motion dynamics: low arousal produces subdued, minimal motion; high arousal increases speed, amplitude, and reach. This variation creates visible difference in energy and expressiveness aligned with emotional intent. To enhance realism, a sinusoidal oscillation is applied to joint angles to simulate breathing. Oscillation intensity scales with arousal, adding subtle motion that avoids stillness and improves lifelikeness. When secondary cues, such as raised hands, are detected, the robot briefly shifts gaze before returning to the primary target, reinforcing a sense of reactivity. In idle states, it orients toward virtual target within reach, maintaining continuity in motion even without human input. The result is a platform-agnostic system for generating socially meaningful gaze using arm motion, modulated by internal state and real-time cues from the environment. 

\subsection{Attentional Drift Module}
Early pilot studies revealed that continuous fixation on a user often felt repetitive or overly mechanical, an observation supported by prior findings showing that excessive gaze persistence can reduce perceived naturalness and engagement in social robots ~\cite{park_towards_2024}. To address this, we introduce an \textit{Attentional Drift Module} that injects naturalistic variability into the robot's gaze behaviour. Inspired by patterns of spontaneous attentional shifts in humans ~\cite{dalmaso_social_2020}, the module generates transient virtual targets within the robot's workspace. These targets are instantiated probabilistically, with their likelihood increasing alongside the robot's arousal level. Each target is assigned a randomized position and brief lifespan, temporarily overriding the robot's primary gaze and prompting a brief redirection before fading. By incorporating these momentary shifts into the attention system, the robot avoids mechanical stiffness, maintains expressive continuity, and fosters a more dynamic, lifelike presence. 

\section{User Study}
\label{sec:user_study}
To evaluate the effectiveness of the proposed behaviour generation system, we conducted a user study investigating how variations in \textit{arousal} and \textit{attention} influence human perceptions of a non-anthropomorphic robot. Building on the system described in Section~\ref{sec:system_design}, this study examines the social and emotional impact of robot behaviour through open-ended human-robot interactions and was approved by the Queen's University General Research Ethics Board (GELEC-139-22, File No. 6036728). 

We specifically explore two behavioural dimensions: \textbf{arousal}, operationalized as the robot's movement energy, and \textbf{attention}, defined as the robot's capacity to direct gaze and responsiveness toward users. These factors were selected due to their central role in shaping user perceptions of social agents, as discussed in Section~\ref{rel-wrk}. A $2 \times 2$ factorial design was employed, crossing two levels of arousal (low and high) with two levels of attention (low and high), resulting in four experimental conditions.

\subsection{Conditions}
\begin{itemize}
    \item \textbf{Low Attention:} The robot remains in an idle state and does not track or respond to human users. It intermittently shifts its gaze toward virtual stimuli generated by the virtual object manager.
    \item \textbf{High Attention:} The robot actively tracks and responds to the participant with gaze and posture adjustments, following the mechanisms described in Sections~\ref{subsec:attn_eng} and~\ref{subsec:gaze_tracking}.
    \item \textbf{Low Arousal:} Arousal is set to its minimum value (1), resulting in slow, conservative motion, a hunched posture, and limited range of movement to create a subdued and deliberate behavioural profile.
    \item \textbf{High Arousal:} Arousal is set to its maximum value (10), producing faster, more dynamic movement with an upright posture and extended reach, conveying a heightened energy level.
\end{itemize}



\subsection{Protocol} 
The study was conducted in a controlled lab environment. Participants first completed a pre-study survey (GAToRS) \cite{koverola_general_2022} assessing baseline attitudes towards robots. Each participant then experiences four interactions sessions, one per experimental condition, with the order counterbalanced using a Latin square to control for carryover effects. 

Participants were instructed to engage with the robot freely, without a specific task or time limit, and to avoid entering to robot's workspace. This open-ended format was chosen to encourage spontaneous behaviour and allow natural social interpretations of the robot's actions. While this approach avoids task-driven bias and supports explorations of perceived social intent, it introduces variability in interaction style and duration across participants. Future comparisons with goal-directed tasks could help contextualize these responses. 

After each session, participants completed a combined post-condition survey (Robotic Social Attributes Scale and Human-Robot Interaction Evaluation Scale) \cite{carpinella_robotic_2017}, \cite{spatola_perception_2021} measuring perceptions of warmth, competence, sociability, animacy, intentionality, discomfort, and disturbance. Participants rated each item on the inventories on scales of one to seven where one represents strongly disagree and seven represents strongly agree. Interactions were audio/video recorded for later analysis, though data from two participants were excluded due to lack of consent or technical error.  

\section{Results}
A total of 36 participants were recruited to participate without compensation through on- and off-campus advertising. The sample included 18 males and 18 females (none identified as non-binary, transgender, or other gender identities), with ages ranging from 18 to 25 years ($M = 21.7$, $SD = 1.58$).

\subsection{Baseline Attitudes}

Pre-study attitudes, as measured by the GAToRS survey~\cite{koverola_general_2022}, revealed generally positive perceptions of robots. On the personal level, participants reported moderate comfort and enjoyment around robots ($P^+ = 4.44$, $SD = 1.52$) and low levels of unease ($P^- = 2.93$, $SD = 1.67$). On the societal level, participants expressed optimism about the impact of robots ($S^+ = 5.92$, $SD = 1.05$), though concerns about potential risks remained present ($S^- = 5.11$, $SD = 1.60$), with more variability in responses.


\subsection{Main Effects and Interaction Analysis}
A $2 \times 2$ within-subjects repeated-measures ANOVA revealed several significant main effects of attention. Attention significantly influenced warmth, $F(1, 35) = 16.97$, $p < .001$, $\eta^2 = .327$; competence, $F(1, 35) = 175.35$, $p < .001$, $\eta^2 = .834$; sociability, $F(1, 35) = 7.22$, $p < .01$, $\eta^2 = .171$; animacy, $F(1, 35) = 18.05$, $p < .001$, $\eta^2 = .340$; and intentionality, $F(1, 35) = 65.32$, $p < .001$, $\eta^2 = .651$.

Arousal also produced significant effects, increasing discomfort, $F(1, 35) = 10.70$, $p < .01$, $\eta^2 = .234$; decreasing sociability, $F(1, 35) = 4.75$, $p < .05$, $\eta^2 = .120$; and increasing disturbance, $F(1, 35) = 7.58$, $p < .01$, $\eta^2 = .178$.

In addition, there were significant Arousal $\times$ Attention interaction effects for discomfort, $F(1, 35) = 4.64$, $p < .05$, $\eta^2 = .117$, and sociability, $F(1, 35) = 5.38$, $p < .05$, $\eta^2 = .133$, indicating that the influence of arousal on these measures depended on the level of attention.

These omnibus effects were followed up with Bonferroni-corrected pairwise comparisons to examine the specific differences between conditions.

\subsection{Post Hoc Comparisons}
Bonferroni-corrected pairwise comparisons revealed the following effects ratings out of seven:

\begin{itemize}
    \item \textbf{Disturbance:} Higher in high ($M = 3.87$, $SD = 0.24$) vs. low arousal ($M = 3.34$, $SD = 0.22$), $p < .01$.
    \item \textbf{Warmth:} Higher in high ($M = 3.77$, $SD = 0.17$) vs. low attention ($M = 3.01$, $SD = 0.15$), $p < .001$.
    \item \textbf{Competence:} Higher in high ($M = 5.11$, $SD = 0.11$) vs. low attention ($M = 3.06$, $SD = 0.17$), $p < .001$.
    \item \textbf{Animacy:} Higher in high ($M = 4.10$, $SD = 0.17$) vs. low attention ($M = 3.44$, $SD = 0.19$), $p < .001$.
    \item \textbf{Intentionality:} Higher in high ($M = 4.95$, $SD = 0.13$) vs. low attention ($M = 3.53$, $SD = 0.16$), $p < .001$.
\end{itemize}

\subsubsection*{Interaction Effects}  
\begin{figure}[ht!]
    \centering
    \includegraphics[width=1.0\linewidth]{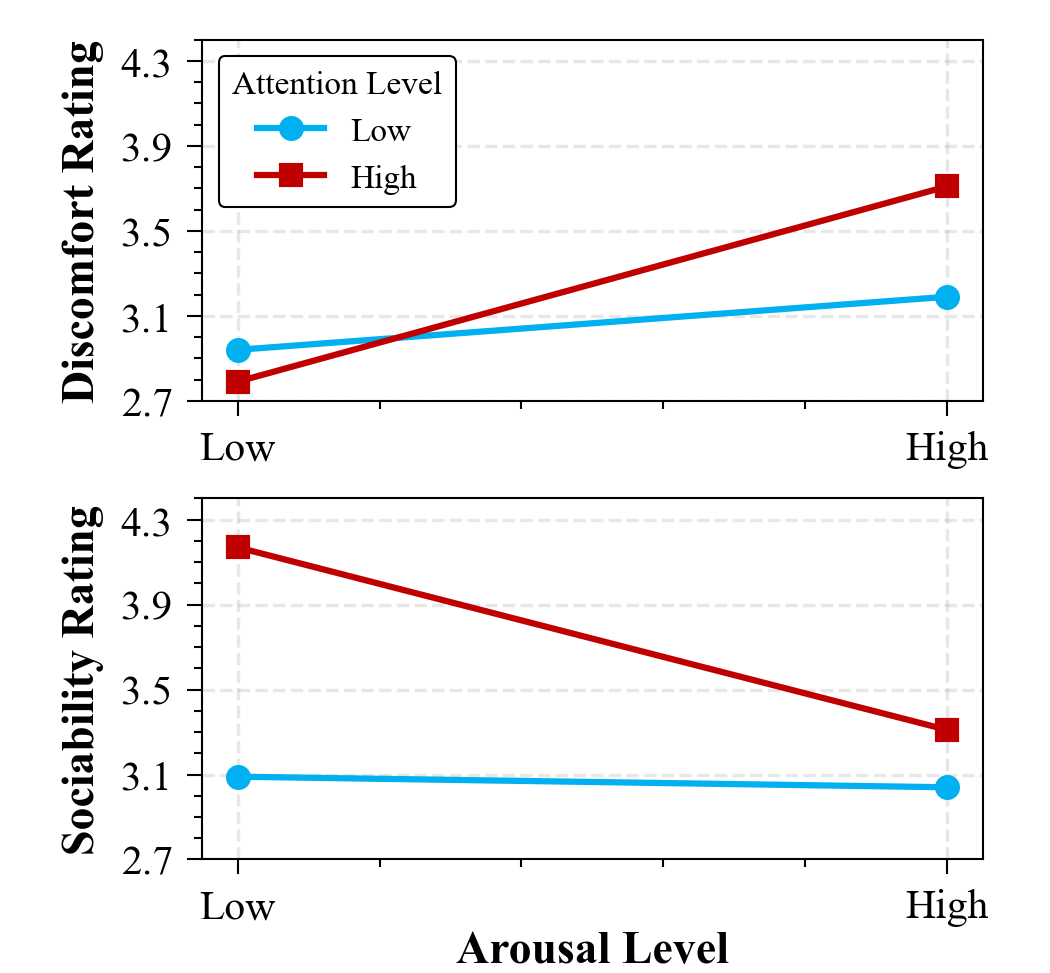}
    \caption{Interaction effect of arousal and attention on discomfort and sociability ratings.}
    \label{fig:int-effs}
\end{figure}

Significant interaction effects are visualized in Fig.~\ref{fig:int-effs}.
In high attention conditions, discomfort ratings were significantly higher in high arousal ($M = 3.71$, $SD = 0.27$) than low arousal ($M = 2.79$, $SD = 0.20$), $p < .01$.
Furthermore, in high attention conditions, sociability ratings were significantly lower in high arousal ($M = 3.31$, $SD = 0.27$) than low arousal ($M = 4.17$, $SD = 0.21$), $p < .01$.

\begin{table}[h]
    \centering
    \renewcommand{\arraystretch}{1.0}
    \caption{Significant Results for Arousal and Attention Effects}
    \label{tab:anova_significant}
    \resizebox{1\linewidth}{!}{
    \begin{tabular}{@{}llccc@{}}
        \toprule
        \textbf{Source} & \textbf{Measure} & \textbf{F} & \textbf{Partial $\eta^2$} & \textbf{Power} \\
        \midrule
        \multirow{3}{*}{Arousal} 
            & Discomfort    & 10.703**  & .234 & .889 \\
            & Sociability   & 4.753*    & .120 & .564 \\
            & Disturbance   & 7.578**   & .178 & .763 \\
        \midrule
        \multirow{5}{*}{Attention} 
            & Warmth        & 16.971*** & .327 & .980 \\
            & Competence    & 175.349***& .834 & 1.000 \\
            & Sociability   & 7.217**   & .171 & .743 \\
            & Animacy       & 18.051*** & .340 & .985 \\
            & Intentionality& 65.315*** & .651 & 1.000 \\
        \midrule
        \multirow{2}{*}{Arousal × Attention} 
            & Discomfort    & 4.637*    & .117 & .553 \\
            & Sociability   & 5.376*    & .133 & .616 \\
        \bottomrule
        \multicolumn{5}{r}{\footnotesize *$p<.05$, **$p<.01$, ***$p<.001$}
    \end{tabular}
    }
\end{table}

\subsection{Interaction Time}

Interaction duration (in seconds) for each trial was also analyzed. Average times by condition were:

\begin{itemize}
    \item Low Arousal, Low Attention: $M = 75.70$, $SD = 33.51$
    \item Low Arousal, High Attention: $M = 68.20$, $SD = 32.23$
    \item High Arousal, Low Attention: $M = 62.97$, $SD = 26.50$
    \item High Arousal, High Attention: $M = 61.44$, $SD = 38.34$
\end{itemize}

Arousal level had a significant effect on interaction duration ($F = 4.66$, $p < .05$, $\eta^2 = .124$), with longer interactions observed in low arousal conditions ($M = 71.96$, $SD = 4.83$) than in high arousal conditions ($M = 62.21$, $SD = 4.99$).


\section{Discussion}
While this study yielded a broad set of findings, we focus here on the three research questions posed earlier: (1) how can NARs convey emotional character through movement, (2) which parameters most influence perceptions of emotional and social engagement, and (3) how do arousal and attention interact to shape impressions of a robot's presence and intent. We structure this discussion around three key themes: expressive attention, arousal and discomfort, and behavioural balance. 

Attention had the most consistent and positive effect on perceived intentionality, warmth, and competence. Even without anthropomorphic features, dynamic orientation toward the user appeared sufficient to suggest social awareness and intent. This aligns with prior work showing that gaze-like behaviour can elicit engagement and perceptions of agency, even in non-humanoid systems \cite{admoni_social_2017}, \cite{moon_meet_2014}. Participants appeared to interpret attentional cues as responsive and deliberate rather than purely mechanical. This supports existing theories of joint attention as a potent expressive  tool, even in minimal or industrial robot designs. For NARs, dynamic orientation may offer a viable alternative to anthropomorphic expressivity, conveying presence without relying on form. 

In contrast, arousal broadly had negative effects on social perception. As arousal increased, participants reported higher levels of discomfort and disturbance, and rated the robot as less sociable. These findings challenge the assumption that increased energy contributes positively to perceived engagement. Instead, excessive motion speed or amplitude may signal volatility, unpredictability, or threat. Prior work show that motion legibility improves trust when movements are predictable and well-scaled to context. In our study, the high-arousal conditions may have violated those expectations, especially in a form factor typically associated with precision and stability. Without anthropomorphic cues to contextualize high-energy behaviour, users may default to interpreting it as chaotic or unsafe. For NARs, this places an upper bound on expressivity through arousal alone: too much motion energy can erode social comfort, even when paired with high attention. 

The most socially successful behaviours emerged when attention was high and arousal remained low. This pairing was was consistently rated as the most sociable, least disturbing, and among the most competent. Users appear to prefer systems that are attentive without being overstimulating. This reflects broader human norms around affect regulation, where calm focus is associated with trustworthiness and emotional control. These findings reinorce principles from frameworks like ELEGNT \cite{hu_elegnt_2025}, which argue that expressive movement should enhance rather than distract from the interaction. For NARs, behaviour must strike a balance between clarity and intensity. Our results suggest that subtlety, rather than raw energy, is key  to sustaining positive social dynamics in robotic arms. 

While this study offers valuable insights into expressive behaviour design for NARs, several limitations warrant consideration. First, interactions were short-term and took place in a controlled laboratory setting with a relatively homogeneous participant pool, that is university students aged 18 to 25. These constraints limit the ecological validity and generalizability of our findings to more divers and real-world contexts. Second, the interactions were intentionally open-ended and task free to encourage spontaneous social interpretations and reduce task-induced bias. However, we recognize that task context likely shapes how users perceive arousal and attention. Third, we employed a fixed, linear arousal parameter to enable clear experimental control. While this simplification facilitated isolating the effects of arousal, it lacks adaptability. Future research could explore dynamic arousal models, such as bio=inspired or homeostatic frameworks that better reflect contextual or user-driven changes.

\section{Conclusion}
This work presents a behaviour generation system for NARs that leverages arousal and attention to support emotionally expressive and socially engaging interactions. Through a controlled user study, we examined how variations in these two dimensions influence human impressions of a robotic manipulator. Our findings show that high attention significantly enhances perceptions of warmth, competence, intentionality, and animacy, while high arousal—expressed through fast and expansive motion—can increase discomfort and disturbance. These results underscore the importance of balancing energy and engagement: attentional cues are effective in eliciting positive social responses, but excessive arousal may diminish the robot’s social acceptability.

The study demonstrates that even robots without human-like features can convey character-like behaviour when designed with appropriate expressive strategies. These insights offer concrete guidance for the design of socially intuitive robotic systems, particularly in non-humanoid forms. While our findings are limited by factors such as the short-term nature of the interactions and scope, this work lays the groundwork for future studies on long-term engagement, adaptive behaviours, and deployment in diverse real-world contexts.


\bibliographystyle{IEEEtran}
\bibliography{cleaned_references}

\begin{thebibliography}{10}
\providecommand{\url}[1]{#1}
\csname url@samestyle\endcsname
\providecommand{\newblock}{\relax}
\providecommand{\bibinfo}[2]{#2}
\providecommand{\BIBentrySTDinterwordspacing}{\spaceskip=0pt\relax}
\providecommand{\BIBentryALTinterwordstretchfactor}{4}
\providecommand{\BIBentryALTinterwordspacing}{\spaceskip=\fontdimen2\font plus
\BIBentryALTinterwordstretchfactor\fontdimen3\font minus \fontdimen4\font\relax}
\providecommand{\BIBforeignlanguage}[2]{{%
\expandafter\ifx\csname l@#1\endcsname\relax
\typeout{** WARNING: IEEEtran.bst: No hyphenation pattern has been}%
\typeout{** loaded for the language `#1'. Using the pattern for}%
\typeout{** the default language instead.}%
\else
\language=\csname l@#1\endcsname
\fi
#2}}
\providecommand{\BIBdecl}{\relax}
\BIBdecl

\bibitem{law_interplay_2021}
T.~Law, M.~Chita-Tegmark, and M.~Scheutz, ``\BIBforeignlanguage{en}{The {Interplay} {Between} {Emotional} {Intelligence}, {Trust}, and {Gender} in {Human}–{Robot} {Interaction}},'' \emph{\BIBforeignlanguage{en}{Int. J. Social Robotics}}, vol.~13, no.~2, pp. 297--309, Apr. 2021.

\bibitem{szafir_pay_2012}
D.~Szafir and B.~Mutlu, ``\BIBforeignlanguage{en}{Pay attention!: designing adaptive agents that monitor and improve user engagement},'' in \emph{\BIBforeignlanguage{en}{Proc. {SIGCHI} Conf. Hum. Factors Comput. Syst.}}\hskip 1em plus 0.5em minus 0.4em\relax Austin Texas USA: ACM, May 2012, pp. 11--20.

\bibitem{bickmore_establishing_2005}
T.~W. Bickmore and R.~W. Picard, ``\BIBforeignlanguage{en}{Establishing and maintaining long-term human-computer relationships},'' \emph{\BIBforeignlanguage{en}{ACM Trans. Comput.-Hum. Interact.}}, vol.~12, no.~2, pp. 293--327, Jun. 2005.

\bibitem{venture_robot_2019}
G.~Venture and D.~Kuli\'{c}, ``Robot {Expressive} {Motions}: {A} {Survey} of {Generation} and {Evaluation} {Methods},'' \emph{J. Hum.-Robot Interact.}, vol.~8, no.~4, pp. 20:1--20:17, Nov. 2019.

\bibitem{hoffman_designing_2014}
G.~Hoffman and W.~Ju, ``Designing robots with movement in mind,'' \emph{J. Hum.-Robot Interact.}, vol.~3, no.~1, pp. 91--122, Feb. 2014.

\bibitem{pan_realistic_2020}
M.~K. Pan, S.~Choi, J.~Kennedy, K.~McIntosh, D.~C. Zamora, G.~Niemeyer, J.~Kim, A.~Wieland, and D.~Christensen, ``Realistic and {Interactive} {Robot} {Gaze},'' in \emph{2020 {IEEE}/{RSJ} Int. Conf. Intell. Robots Syst. ({IROS})}, Oct. 2020, pp. 11\,072--11\,078, iSSN: 2153-0866.

\bibitem{spatola_ascribing_2021}
N.~Spatola and O.~A. Wudarczyk, ``Ascribing emotions to robots: {Explicit} and implicit attribution of emotions and perceived robot anthropomorphism,'' \emph{Comput. Hum. Behav.}, vol. 124, p. 106934, Nov. 2021.

\bibitem{chevalier_dialogue_2017}
P.~Chevalier, J.~J. Li, E.~Ainger, A.~M. Alcorn, S.~Babovic, V.~Charisi, S.~Petrovic, B.~R. Schadenberg, E.~Pellicano, and V.~Evers, ``\BIBforeignlanguage{en}{Dialogue {Design} for a {Robot}-{Based} {Face}-{Mirroring} {Game} to {Engage} {Autistic} {Children} with {Emotional} {Expressions}},'' in \emph{\BIBforeignlanguage{en}{Social Robotics}}, A.~Kheddar, E.~Yoshida, S.~S. Ge, K.~Suzuki, J.-J. Cabibihan, F.~Eyssel, and H.~He, Eds.\hskip 1em plus 0.5em minus 0.4em\relax Cham: Springer Int. Publ., 2017, pp. 546--555.

\bibitem{kim_how_2014}
Y.~Kim and B.~Mutlu, ``How social distance shapes human–robot interaction,'' \emph{Int. J. Hum.-Comput. Stud.}, vol.~72, no.~12, pp. 783--795, Dec. 2014.

\bibitem{green_whos_2022}
H.~N. Green, M.~M. Islam, S.~Ali, and T.~Iqbal, ``Who's {Laughing} {NAO}? {Examining} {Perceptions} of {Failure} in a {Humorous} {Robot} {Partner},'' in \emph{Proc. 2022 {ACM}/{IEEE} Int. Conf. Hum.-Robot Interact.}, ser. {HRI} '22.\hskip 1em plus 0.5em minus 0.4em\relax Sapporo, Hokkaido, Japan: IEEE Press, Mar. 2022, pp. 313--322.

\bibitem{saunderson_how_2019}
S.~Saunderson and G.~Nejat, ``How robots influence humans: {A} survey of nonverbal communication in social human–robot interaction,'' \emph{Int. J. Social Robotics}, vol.~11, no.~4, pp. 575--608, 2019.

\bibitem{stanton_robot_2014}
C.~Stanton and C.~Joanna~Stevens, ``Robot {Pressure}: {The} {Impact} of {Robot} {Eye} {Gaze} and {Lifelike} {Bodily} {Movements} upon {Decision}-{Making} and {Trust},'' \emph{6th Int. Conf. Social Robotics ICSR}, Oct. 2014.

\bibitem{moon_meet_2014}
A.~J. Moon, D.~M. Troniak, B.~Gleeson, M.~K. Pan, M.~Zheng, B.~A. Blumer, K.~MacLean, and E.~A. Croft, ``Meet {Me} where {I}'m {Gazing}: {How} {Shared} {Attention} {Gaze} {Affects} {Human}-{Robot} {Handover} {Timing},'' in \emph{2014 9th {ACM}/{IEEE} Int. Conf. Hum.-Robot Interact. ({HRI})}, Mar. 2014, pp. 334--341, iSSN: 2167-2121.

\bibitem{mori_uncanny_2012}
M.~Mori, K.~F. MacDorman, and N.~Kageki, ``The {Uncanny} {Valley} [{From} the {Field}],'' \emph{IEEE Robot. Autom. Mag.}, vol.~19, no.~2, pp. 98--100, Jun. 2012.

\bibitem{kim_robotic_2022}
L.~H. Kim, V.~Domova, Y.~Yao, C.-M. Huang, S.~Follmer, and P.~E. Paredes, ``Robotic {Presence}: {The} {Effects} of {Anthropomorphism} and {Robot} {State} on {Task} {Performance} and {Emotion},'' \emph{IEEE Robot. Autom. Lett.}, vol.~7, no.~3, pp. 7399--7406, Jul. 2022.

\bibitem{scheflen_significance_1964}
A.~E. Scheflen, ``\BIBforeignlanguage{eng}{The significance of posture in communication systems},'' \emph{\BIBforeignlanguage{eng}{Psychiatry}}, vol.~27, pp. 316--331, Nov. 1964.

\bibitem{erden_emotional_2013}
M.~S. Erden, ``\BIBforeignlanguage{en}{Emotional {Postures} for the {Humanoid}-{Robot} {Nao}},'' \emph{\BIBforeignlanguage{en}{Int. J. Social Robotics}}, vol.~5, no.~4, pp. 441--456, Nov. 2013.

\bibitem{zecca_whole_2009}
M.~Zecca, Y.~Mizoguchi, K.~Endo, F.~Iida, Y.~Kawabata, N.~Endo, K.~Itoh, and A.~Takanishi, ``Whole body emotion expressions for {KOBIAN} humanoid robot — preliminary experiments with different {Emotional} patterns —,'' in \emph{{RO}-{MAN} 2009 - 18th {IEEE} Int. Symp. Robot Hum. Interact. Commun.}, Sep. 2009, pp. 381--386, iSSN: 1944-9437.

\bibitem{takayama_expressing_2011}
L.~Takayama, D.~Dooley, and W.~Ju, ``Expressing thought: {Improving} robot readability with animation principles,'' in \emph{2011 6th {ACM}/{IEEE} Int. Conf. Hum.-Robot Interact. ({HRI})}, Mar. 2011, pp. 69--76, iSSN: 2167-2148.

\bibitem{hu_elegnt_2025}
Y.~Hu, P.~Huang, M.~Sivapurapu, and J.~Zhang, ``{ELEGNT}: {Expressive} and {Functional} {Movement} {Design} for {Non}-anthropomorphic {Robot},'' Jan. 2025, arXiv:2501.12493 [cs].

\bibitem{breazeal_emotion_2003}
C.~Breazeal, ``Emotion and sociable humanoid robots,'' \emph{Int. J. Hum.-Comput. Stud.}, vol.~59, no.~1, pp. 119--155, Jul. 2003.

\bibitem{russell_circumplex_1980}
J.~A. Russell, ``\BIBforeignlanguage{en}{A {Circumplex} {Model} of {Affect}},'' \emph{\BIBforeignlanguage{en}{Journal of Personality and Social Psychology}}, Dec. 1980.

\bibitem{admoni_social_2017}
H.~Admoni and B.~Scassellati, ``Social {Eye} {Gaze} in {Human}-{Robot} {Interaction}: {A} {Review},'' \emph{J. Hum.-Robot Interact.}, vol.~6, no.~1, p.~25, Mar. 2017.

\bibitem{noauthor_ur5e_nodate}
\BIBentryALTinterwordspacing
``{UR5e} {Lightweight}, versatile cobot.'' [Online]. Available: \url{https://www.universal-robots.com/products/ur5-robot/}
\BIBentrySTDinterwordspacing

\bibitem{noauthor_intel_nodate}
``\BIBforeignlanguage{en}{Intel® {RealSense}™ {Depth} {Camera} {D435} - {Product} {Specifications}}.''

\bibitem{park_towards_2024}
J.~Park, T.~Jeong, H.~Kim, T.~Byun, S.~Shin, K.~Choi, J.~Kwon, T.~Lee, M.~Pan, and S.~Choi, ``Towards {Embedding} {Dynamic} {Personas} in {Interactive} {Robots}: {Masquerading} {Animated} {Social} {Kinematic} ({MASK}),'' \emph{IEEE Robot. Autom. Lett.}, vol.~9, no.~10, pp. 8826--8833, Oct. 2024, conf. Name: IEEE Robot. Autom. Lett.

\bibitem{dalmaso_social_2020}
M.~Dalmaso, L.~Castelli, and G.~Galfano, ``\BIBforeignlanguage{en}{Social modulators of gaze-mediated orienting of attention: {A} review},'' \emph{\BIBforeignlanguage{en}{Psychon. Bull. Rev.}}, vol.~27, no.~5, pp. 833--855, Oct. 2020.

\bibitem{koverola_general_2022}
M.~Koverola, A.~Kunnari, J.~Sundvall, and M.~Laakasuo, ``General {Attitudes} {Towards} {Robots} {Scale} ({GAToRS}): {A} new instrument for social surveys,'' \emph{Int. J. Social Robotics}, vol.~14, no.~7, pp. 1559--1581, 2022.

\bibitem{carpinella_robotic_2017}
C.~M. Carpinella, A.~B. Wyman, M.~A. Perez, and S.~J. Stroessner, ``The {Robotic} {Social} {Attributes} {Scale} ({RoSAS}): {Development} and {Validation},'' in \emph{2017 12th {ACM}/{IEEE} Int. Conf. Hum.-Robot Interact. ({HRI})}, Mar. 2017, pp. 254--262, iSSN: 2167-2148.

\bibitem{spatola_perception_2021}
N.~Spatola, B.~Kuhnlenz, and G.~Cheng, ``Perception and {Evaluation} in {Human}–{Robot} {Interaction}: {The} {Human}–{Robot} {Interaction} {Evaluation} {Scale} ({HRIES})—{A} {Multicomponent} {Approach} of {Anthropomorphism},'' in \emph{Int. J. Social Robotics}, vol.~13, Jan. 2021, pp. 1517--1539.

\end{thebibliography}
\end{document}